\pdfoutput=1
\documentclass[11pt]{article}
\usepackage{arabtex}
\usepackage[final]{acl}

\usepackage{times}
\usepackage{latexsym}
\usepackage{fontawesome}
\usepackage[T1]{fontenc}
\usepackage[utf8]{inputenc}

\usepackage{microtype}

\usepackage{inconsolata}

\usepackage{graphicx}

\usepackage{booktabs} % For professional table lines
\usepackage{graphicx} % For including graphics
\usepackage{xcolor}   % For coloring rows or cells
\usepackage{amssymb}  % For checkmark symbol
\usepackage{array}    % For advanced column formatting

\usepackage{amsmath}
\usepackage{amsfonts}
\usepackage{geometry}
\usepackage{cuted} % Add to preamble
\usepackage{caption} % Add to preamble
\usepackage{lipsum}

\usepackage[most]{tcolorbox}

\usepackage{rotating}

\usepackage{multirow}
\usepackage{multicol}

\usepackage{colortbl}
\usepackage{adjustbox}

\definecolor{mainheader}{RGB}{233, 236, 239}
\definecolor{subheader}{RGB}{241, 243, 244}
\definecolor{teamname}{RGB}{255, 243, 205}

\newcolumntype{L}[1]{>{\raggedright\arraybackslash}p{#1}}
\newcolumntype{C}[1]{>{\centering\arraybackslash}p{#1}}

\usepackage{utf8}
\setcode{utf8}

\usepackage[utf8]{inputenc}
\usepackage[T2A,LAE,T1]{fontenc}
\usepackage[arabic,USenglish]{babel}
\usepackage[titletoc]{appendix}

\title{
\raisebox{-2.1ex}{\protect\includegraphics[height=3.99\fontcharht\font`\B]{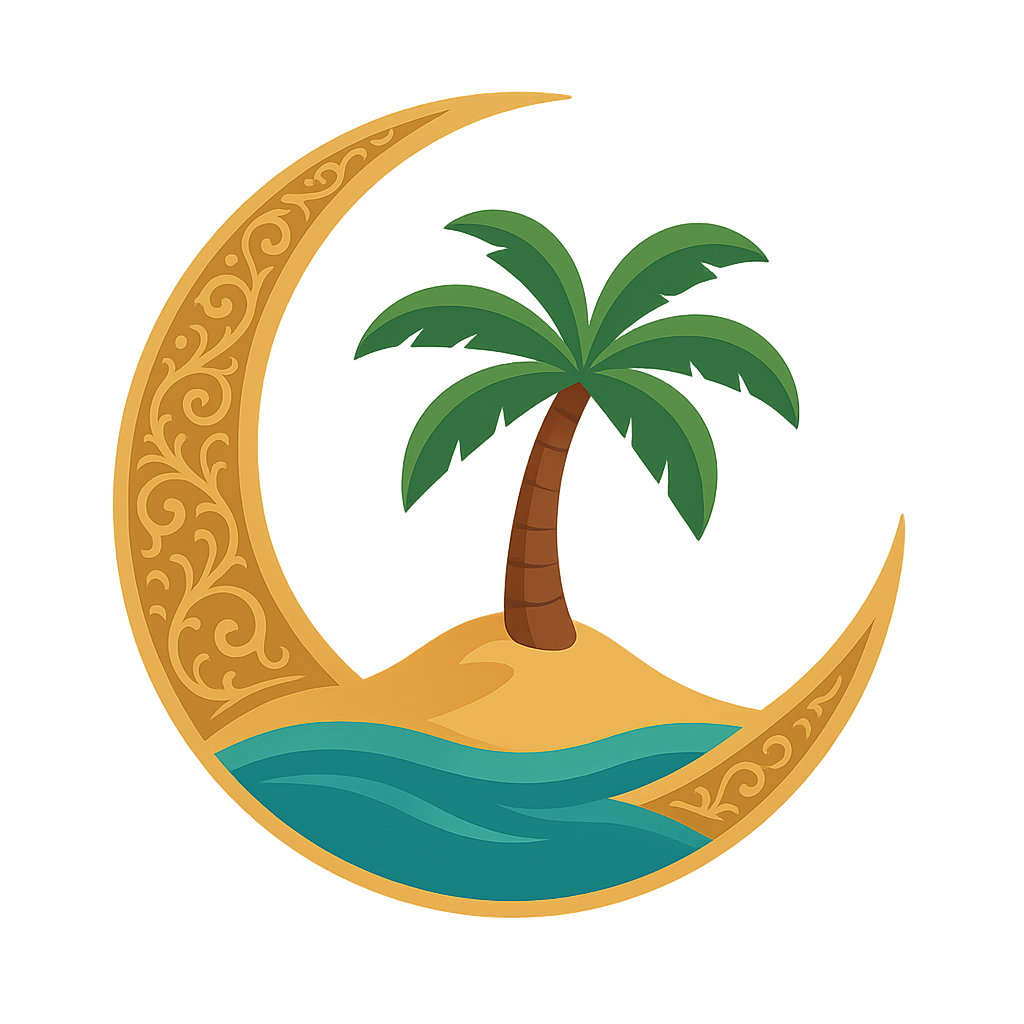}} PalmX 2025: The First Shared Task on Benchmarking LLMs on Arabic and Islamic Culture}

\author{
    \normalsize \bfseries Fakhraddin Alwajih$^{\lambda}$ ~~~~~~~~
    \bfseries Abdellah {El Mekki}$^{\lambda}$ ~~~~~~~~
    \bfseries Hamdy Mubarak$^{\gamma}$ \\ 
    \normalsize \bfseries Majd Hawasly$^{\gamma}$ ~~~~~~~~ 
    \normalsize \bfseries Abubakr Mohamed$^{\gamma}$ ~~~~~~~~ 
    \bfseries Muhammad Abdul-Mageed$^{\lambda}$ \\[1ex] 
    $^{\lambda}$The University of British Columbia ~~~~~~~~~
    $^{\gamma}$Qatar Computing Research Institute \\[1ex]
    \texttt{ \{fakhr.alwajih,abdellah.elmekki,muhammad.mageed\}@ubc.ca} \\
    \texttt{ \{hmubarak,mhawasly,abumohamed\}@hbku.edu.qa} \\
}

\begin{document}
\maketitle

\begin{strip}
    \centering
    \includegraphics[width=0.95\linewidth]{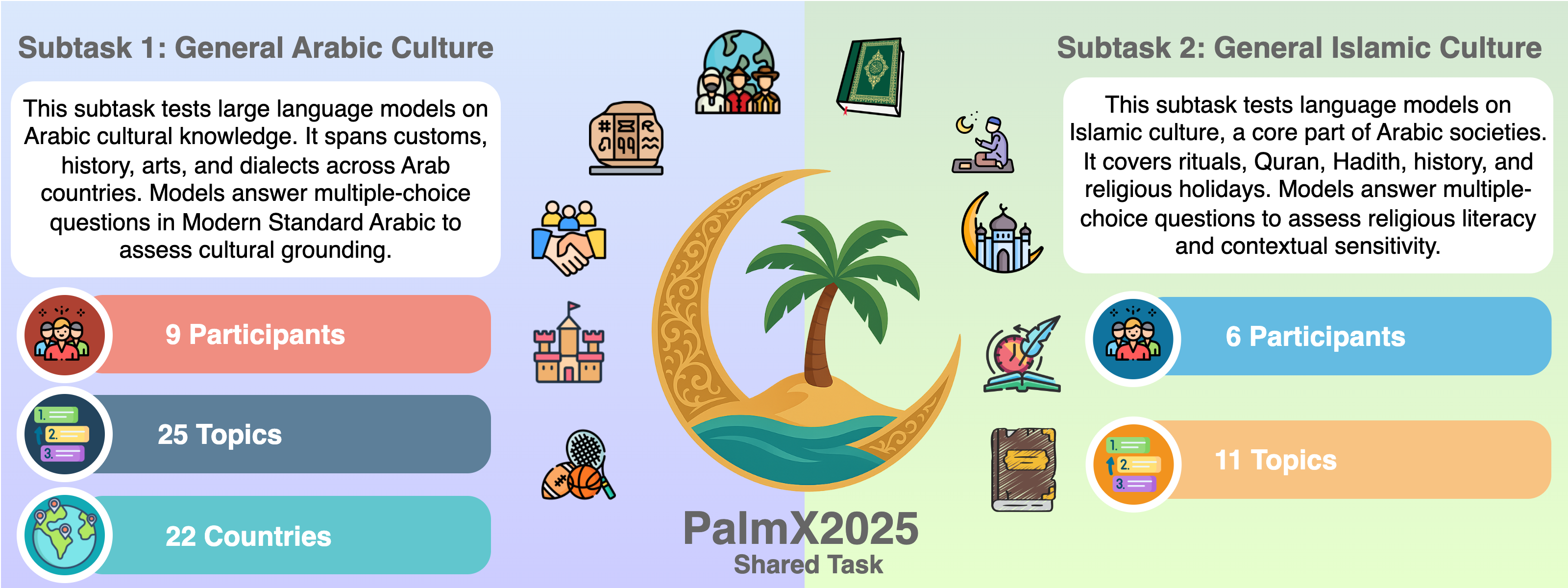}
    \captionof{figure}{An Overview of the PalmX 2025 Shared Task}
    \label{fig:placeholder}
\end{strip}

\begin{abstract}

 Large Language Models (LLMs) inherently reflect the vast data distributions they encounter during their pre-training phase. As this data is predominantly sourced from the web, there is a high chance it will be skewed towards high-resourced languages and cultures, such as those of the West. Consequently, LLMs often exhibit a diminished understanding of certain communities, a gap that is particularly evident in their knowledge of Arabic and Islamic cultures. This issue becomes even more pronounced with increasingly under-represented topics. To address this critical challenge, we introduce PalmX~2025, the first shared task designed to benchmark the cultural competence of LLMs in these specific domains. The task is composed of two subtasks featuring multiple-choice questions (MCQs) in Modern Standard Arabic (MSA): General Arabic Culture and General Islamic Culture. These subtasks cover a wide range of topics, including traditions, food, history, religious practices, and language expressions from across 22 Arab countries. The initiative drew considerable interest, with 26 teams registering for Subtask~1 and 19 for Subtask~2, culminating in nine and six valid submissions, respectively. Our findings reveal that task-specific fine-tuning substantially boosts performance over baseline models. The top-performing systems achieved an accuracy of 72.15\% on cultural questions and 84.22\% on Islamic knowledge. Parameter-efficient fine-tuning emerged as the predominant and most effective approach among participants, while the utility of data augmentation was found to be domain-dependent. Ultimately, this benchmark provides a crucial, standardized framework to guide the development of more culturally grounded and competent Arabic LLMs. Results of the shared task demonstrate that general cultural and general religious knowledge remain challenging to LLMs, motivating us to continue to offer the shared task in the future. 

%Large Language Models (LLMs) often exhibit biases towards Western cultures, leading to inappropriate outputs in Arabic contexts. We introduce PalmX 2025, the first shared task benchmarking LLMs on Arabic and Islamic cultural competence. The task features two subtasks with multiple-choice questions in Modern Standard Arabic: General Arabic Culture and General Islamic Culture, covering traditions, history, religious practices, and dialectal expressions across 22 Arab countries. With 26 teams registering for Subtask 1 and 19 for Subtask 2, we received 9 and 6 valid submissions respectively. Task-specific fine-tuning significantly improved performance over baselines, with top performers achieving 72.15\% on cultural questions and 84.22\% on Islamic knowledge. Parameter-efficient fine-tuning (LoRA) emerged as the dominant approach, while data augmentation showed domain-dependent effectiveness. This benchmark provides a standardized framework for developing culturally grounded Arabic LLMs.
\end{abstract}

\section{Introduction}
\label{sec:introduction}

Despite their impressive capabilities, LLMs often display systematic Western- and Anglocentric biases, mirroring the over-representation of these perspectives in their training data~\cite{adilazuarda2024towards,pawar2025survey}. This lack of cultural diversity can lead to outputs that are not only inappropriate but also harmful. For instance, an Arabic LLM trained on translated English data once suggested having a beer after prayer, a recommendation that fundamentally misunderstands (and indeed disrespects) core Arab cultural and religious norms~\cite{naous2023having}. Incidents such as this underscore a critical distinction in LLM development between \textit{cultural awareness}, which refers to the understanding of a culture's norms and values and \textit{cultural alignment}, which is focused on the adaptation of actions to respect and reflect these norms and values~\cite{alkhamissi2024investigating}. True progress requires models that are not just culturally aware, but culturally aligned as well.

The need for culturally aligned models is particularly acute in the Arab world, a region of over 450 million people spread across 22 countries. The Arab world comprises immense diversity in customs and traditions, as well as dialectal richness. While recent efforts have produced relatively fluent Arabic LLMs~\cite{bari2024allam,sengupta2023jais,huang-etal-2024-acegpt}, many are trained on machine-translated datasets and evaluated on general NLP tasks in ways that largely overlook country-specific cultural competence. Foundational work on datasets like \textit{Palm}~\cite{alwajih2025palm} has begun to address this by providing culturally inclusive, human-created Arabic instructions covering all 22 Arab countries. However, a standardized benchmark is still needed to systematically measure and compare the cultural understanding of different models.

To bridge this evaluation gap, we introduce the \textit{PalmX~2025} Shared Task, the first benchmarking effort focused specifically on the cultural competence of LLMs in the Arabic context. In this task, we define culture as the collection of knowledge, beliefs, and behaviors encompassing the traditions, social etiquette, cuisine, history, arts, dialectal expressions, and religious practices that characterize communities across the Arab world. \textit{PalmX} challenges models with multiple-choice questions designed to test deep cultural knowledge, not superficial pattern matching. The task is divided into two subtasks: one on \emph{General Arabic Culture} and another on \emph{General Islamic Culture}, reflecting the cornerstones of identity in the region. By providing a standardized evaluation framework, \textit{PalmX} aims to drive the development of LLMs that are not only linguistically fluent but also culturally grounded and respectful.

This paper is organized as follows: Section~\ref{sec:task_desc} describes the \textit{PalmX~2025} shared task, including data collection and annotation for both subtasks. Section~\ref{sec:rules_eval} outlines the participation rules and evaluation methodology. Section~\ref{sec:teams_results} presents the participating teams and their results. Section~\ref{sec:discussion} discusses the findings and provides analysis of the methodological approaches for the participating teams. Section~\ref{sec:conclusion} concludes with key insights and future directions. Appendix~\ref{app:lit_review} provides a literature review of related work, and Appendix~\ref{app:data_analysis} presents detailed data analysis including country and topic distributions for datasets of both subtasks.

\section{Task Description: PalmX 2025}
\label{sec:task_desc}

 The objective of the \textit{PalmX~2025} Shared Task\footnote{\url{https://palmx.dlnlp.ai/}} is to enable evaluation of the competence of LLMs on Arabic and Islamic cultures through two independent subtasks: \textit{general Arabic culture} and \textit{general Islamic culture}. Each subtask is designed as a set of MCQs in MSA, each with four options (A-D) and a single correct answer; the questions target grounded knowledge. The distractors for each MCQ question are designed to plausible but incorrect, often sharing surface cues to minimize the chance of correct guesses. For each subtask, we provide training, development (dev), and test splits. The training split is provided to participants to support system development, allowing for various approaches such as fine-tuning. Additionally, the dev split is shared with participants to facilitate hyperparameter tuning and local evaluation of their systems before the test phase. The test split is kept private during the competition and is released publicly after the competition concludes. We apply basic quality filters to ensure clarity, a single unambiguous answer, and cultural correctness. This process involves removing off-topic questions unrelated to culture, those with multiple correct answers, biased content, and items with grammatical errors. Accuracy is the primary evaluation metric.

%\textcolor{green}{The PalmX 2025} Shared Task~\footnote{\url{https://palmx.dlnlp.ai/}} evaluates how well language models understand Arabic culture across two focused subtasks. Each subtask uses multiple-choice questions (MCQs) in Modern Standard Arabic (MSA). The questions target culturally grounded knowledge, not just surface pattern matching. Systems must select the single correct answer from four options. We provide separate training, development (dev), and blind test splits. Accuracy is the primary evaluation metric.

 All the resources of \textit{PalmX~2025} shared task are publicly available, including data and evaluation code.\footnote{\url{https://github.com/UBC-NLP/palmx_2025}}

\subsection{Subtask 1: General Arabic Culture}
\label{sec:subtask1}

The goal of this subtask is to encourage development of methods for incorporating Arabic general culture in LLMs, allowing them to comprehend and reason about diverse aspects of general Arabic culture. These aspects are coming from different cultural categories including \textit{traditional customs}, \textit{local etiquette}, \textit{cuisine}, \textit{historical events}, \textit{famous figures}, \textit{geography}, \textit{local languages (dialects)}, and \textit{arts}.

\subsubsection{Data Collection and Annotation}\label{subsubsec:data}
The data for this subtask cover a number of cultural topics. To ensure this wide coverage, we follow two complementary data collection strategies, as described below.
%\textcolor{green}{To ensure a balanced coverage} of cultural topics, we employ two complementary data collection strategies as we describe below. \newline

\noindent \textbf{Method 1:} We source the data from \textit{Palm}~\cite{alwajih2025palm}  training split, which we convert into an MCQ format using Qwen3 30B~\cite{yang2025qwen3}. Using this method, we acquire $4,000$ samples.\newline
\noindent \textbf{Method 2:} We crawl web pages from diverse online resources covering cultural knowledge, customs, etiquette, values, and practices across all Arab countries. Representative sources include \textit{Cultural Crossing},{\footnote {\url{https://guide.culturecrossing.net/basics_business_student_details.php}} \textit{Commisceo}\footnote{\url{https://www.commisceo-global.com/resources/country-guides/}}, \textit{Cultural Atlas},\footnote{\url{https://culturalatlas.sbs.com.au/countries}} and \textit{Expatica}.\footnote{\url{https://www.expatica.com/}} We then segment the collected pages into sections and subsections, and employ GPT-4o-mini to generate culturally relevant MCQs in both Arabic and English. We acquire $1,000$ samples using this method.

For both methods, two professional linguists independently reviewed the data for correctness, removal of low-quality or trivial questions, and acquisition of proper formatting. All discrepancies were reviewed in consolidation sessions. Finally, we shuffl answer options to minimize positional bias.

The final data for this subtask consists of $2,000$, $500$, and $2,000$ questions for the training, dev, and test splits, respectively. The domain and country balance in the test set approximates that of the training data but includes some new entities and less frequent cultural items to test generalization.

\noindent
Samples from Subtask 1 are presented in Table \ref{tab:examples_culture}.

\begin{table*}[!ht]
\resizebox{\textwidth}{!}{%
\centering
\small
% We adjusted the 'p' column widths. You may need to tweak these values.
\begin{tabular}{llp{2.5cm}p{2.5cm}p{2.5cm}p{2cm}p{6.5cm}}
\toprule
\textbf{Split} & \textbf{Answer} & \textbf{D} & \textbf{C} & \textbf{B} & \textbf{A} & \textbf{Question} \\
\midrule

train & D & \<23 سبتمبر> & \<1 يناير> & \<14 فبراير> & \<30 نوفمبر> & \<متى يحتفل السعوديون باليوم الوطني؟> \\
train & D & 23 September & 1 January & 14 February & 30 November & When do Saudis celebrate National Day? \\
\addlinespace 
\rowcolor{gray!5}
train & B & \<الصداقة> & \<الحب> & \<الحزن> & \<الفرح> & \<ماذا ترمز زهور البنفسج في الثقافة الجزائرية؟> \\
\rowcolor{gray!5}
train & B & Friendship & Love & Sadness & Joy & What do violets symbolize in Algerian culture? \\
\addlinespace

dev  & D & \<طقس مهم بعد الزفاف> & \<عملية خاصة بالعروس> & \<نوع من الطعام> & \<مباراة تقليدية> & \<ما هو الجرتق في الزواج السوداني؟> \\
dev  & D & An important post-wedding ritual & A special process for the bride & A type of food & A traditional contest & What is "Jertiq" in Sudanese weddings? \\
\addlinespace
\rowcolor{gray!5}
test  & A & \<الفرنسية> & \<البرتغالية> & \<الإيطالية> & \<الأمازيغية> & \<ما هي اللغة الأم لبعض المغاربة بجانب العربية؟> \\
\rowcolor{gray!5}
test  & A & French & Portuguese & Italian & Amazigh & What is the mother tongue of some Moroccans besides Arabic? \\

test  & A & \<المجبوس> & \<الثريد> & \<الكسكس> & \<الكسرة> & \<ما هو الطبق الموريتاني الأكثر شيوعاً في العالم العربي؟> \\
test  & A & Majboos & Thareed & Couscous & Kesra & What is the most common Mauritanian dish in the Arab world? \\
\addlinespace

\bottomrule
\end{tabular}%
}
\caption{Sample questions with their splits, correct answers, and options (A–D) for Subtask 1.
}
\label{tab:examples_culture}
\end{table*}

\subsection{Subtask 2: General Islamic Culture}
\label{sec:subtask2}
This subtask aims to assess the capacity of LLMs to capture and understand the Islamic culture,
%a LLM's understanding of key elements of Islamic culture,
which plays a foundational role in Arabic societies. It covers topics such as \textit{Islamic rituals and practices (e.g., prayers and fasting)}, \textit{Quranic knowledge}, \textit{Hadith literature}, \textit{historical developments in Islam}, and \textit{religious holidays}. %Language Models are expected to answer multiple-choice questions that reflect both religious literacy and contextual sensitivity, ensuring their ability to handle culturally and theologically significant content with accuracy and respect.
\subsubsection{Data Collection and Annotation}
To enhance topical diversity, we employ two complementary methods to collect Islamic MCQs, yielding a nearly balanced distribution across sources.\newline
\noindent
\textbf{Method 1:} We create the data based on public Islamic competitions and general questions about Islamic culture using a university book~\footnote{The Question Bank for Islamic Culture form Al-Balqa Applied University (BAU)}.  We acquire $900$ samples using this method.\newline
\noindent
\textbf{Method 2:} We crawl all Islamic articles from \textit{Mawdoo3},~\footnote{\url{https://mawdoo3.com}} one of the most reputable Arabic content platforms (category: Islam). From this corpus, we randomly select 200 pages and employ GPT-4o-mini to generate diverse MCQs per page. All generated Arabic items are independently reviewed by two professional linguists to verify correctness, eliminate low-quality or trivial content, and ensure proper formatting. Again, all discrepancies are reviewed in consolidation sessions and answer options are subsequently shuffled to reduce positional bias. We acquire $1,000$ samples using this method.

The final data for this subtask consists of $600$, $300$, and $1,000$ questions for the training, dev, and test splits, respectively.

\noindent
Samples from Subtask~2 are presented in Table \ref{tab:examples_islamic}.

\begin{table*}[!ht]
\resizebox{\textwidth}{!}{%
\centering
\small
% We adjusted the 'p' column widths. You may need to tweak these values.
\begin{tabular}{llp{3cm}p{3cm}p{3cm}p{3cm}p{6.5cm}}
\toprule
\textbf{Split} & \textbf{Answer} & \textbf{D} & \textbf{C} & \textbf{B} & \textbf{A} & \textbf{Question} \\
\midrule

dev & B & \<رحمة لا تتعلق بالله> & \<رحمة محدودة> & \<رحمة تشمل جميع \\ المخلوقات> & \<رحمة خاصة \\ بالمؤمنين> & \<أي من العبارات التالية \\ تعبر عن معنى اسم الرحمن؟> \\
dev & B & Mercy unrelated to God & Limited mercy & Mercy that includes all creatures & Mercy specific to believers & Which of the following phrases expresses the meaning of the name "Ar-Rahman"? \\
%\addlinespace
\rowcolor{gray!5}
train & A & \<عثمان بن عفان> & \<معاذ بن جبل> & \<عبد الرحمن بن عوف> & \<أبو عبيدة بن \\ الجراح> & \<من هو الصحابي الذي \\ لُقّب بأمين هذه الأمة؟> \\
\rowcolor{gray!5}
train & A & Uthman ibn Affan & Muadh ibn Jabal & Abdur Rahman ibn Awf & Abu Ubaidah ibn al-Jarrah & Which companion was nicknamed "the trustworthy of this nation"? \\
%\addlinespace

test & D & \<رفيدة بنت سعد \\ الأسلمية رضي الله عنها> & \<حفصة بنت عمر \\ رضي الله عنها> & \<عائشة بنت أبي بكر \\ رضي الله عنها> & \<أم أيمن رضي الله عنها> & \<من هي أول ممرضة \\ في الإسلام؟> \\
test & D & Rufaidah bint Sa’d al-Aslamiyyah (may Allah be pleased with her) & Hafsa bint Umar (may Allah be pleased with her) & Aisha bint Abu Bakr (may Allah be pleased with her) & Umm Ayman (may Allah be pleased with her) & Who was the first nurse in Islam? \\
%\addlinespace

\bottomrule
\end{tabular}%
}
\caption{Sample questions with their splits, correct answers, and options (A–D) for Subtask 2.}
\label{tab:examples_islamic}
\end{table*}

\section{Rules and Evaluation}
\label{sec:rules_eval}
This section outlines the rules we establish for participation and the methods we employ for the evaluation of submissions. We design the framework to rigorously and fairly assess the intrinsic cultural and Islamic knowledge of the submitted language models.

\paragraph{Reproducibility}
Teams are instructed to document their data preprocessing, model architecture, external resources, prompt templates, and inference-time strategies.

\subsection{Participation and Submission Guidelines}
The primary objective of the shared task is to assess the internalized knowledge of LLMs. To ensure the evaluation focuses on the models' core understanding rather than their ability to query external information sources, we established two fundamental rules.

First, the use of systems with real-time data retrieval capabilities, such as retrieval-augmented generation (RAG) or live internet access, is strictly prohibited. This ensures that the task does not become a trivial information retrieval challenge. Consequently, submissions are limited to the following format:
\begin{enumerate}
    \item \textbf{Model Weights:} Participants are required to submit the fine-tuned weights of a decoder-only generative language model.
    \item \textbf{Parameter Limit:} To maintain computational fairness across all participants, the submitted models are constrained to a maximum size of 13 billion ($13$B) parameters.
    \item \textbf{Secure Submission:} For privacy and accessibility, participants are instructed to host their models in a private repository on Hugging Face. The final submission consists of the repository ID and a fine-grained access token that provided the organizers with read-only access to the model for evaluation.
\end{enumerate}

Second, to ensure integrity of the results, the test set was held out and remained private to the organizers\footnote{Test data was  shared only after the leaderboard announcement.}. This blind evaluation protocol guarantees that no participant had prior access to the test data, enabling a realistic assessment of each model's generalization capabilities in the domain of Arabic cultural and Islamic awareness.

\subsection{Evaluation Method}

To evaluate the MCQs  from our test set, we adopt the likelihood-based method commonly used in frameworks like the EleutherAI Language Model Evaluation Harness \cite{biderman2024lessonstrenchesreproducibleevaluation}. This approach assesses a model's understanding by measuring how likely it is to choose the correct answer label after being presented with the question and all possible choices, rather than relying on generative decoding. We develop an in-house script to implement this method and share it with participants during the development phase to ensure they understand how their submissions would be evaluated.

\subsubsection{Likelihood-based MCQ Evaluation}

For each MCQ item, we construct a prompt that includes the question followed by the list of choices, each prefixed with a letter (e.g., A, B, C, D). The prompt is structured as follows:

\begin{tcolorbox}[]
<Question>

A. <Choice 1>

B. <Choice 2>

C. <Choice 3>

D. <Choice 4>

Answer:
\end{tcolorbox}

The model's task is to determine which choice label (A, B, C, or D) is the most probable continuation of the prompt. We calculate the likelihood of the model generating each choice label. This approach of scoring only the label, rather than the full text of the choice, ensures the evaluation is not biased by the length of the answer strings.

Specifically, for a given question prompt $P$ and a set of possible choices $\{C_1, C_2, \dots, C_n\}$, we create $n$ distinct sequences. Each sequence is formed by concatenating the prompt $P$ with the text corresponding to one of the choice labels (e.g., " A", " B", etc.).

Let the tokens for the choice label $C_i$ be $c_{i,1}, c_{i,2}, \dots, c_{i,k}$. The score for choice $C_i$ is its log-likelihood, calculated as the sum of the conditional log-probabilities of its tokens given the prompt and the preceding tokens of the choice label:
\begin{multline}
\text{score}(C_i) = \log p(C_i | P) = \\
\sum_{j=1}^{k} \log p(c_{i,j} | P, c_{i,1}, \dots, c_{i,j-1})
\end{multline}

These log-likelihood scores are computed for all choices. To select the model's final answer, we normalize these scores into a probability distribution using the softmax function:
$$
\text{P}(C_i) = \frac{e^{\text{score}(C_i)}}{\sum_{j=1}^{n} e^{\text{score}(C_j)}}
$$

The choice with the highest resulting probability is selected as the model's prediction.

\subsubsection{Evaluation Metric}

The final performance is measured using \textbf{accuracy}. The model's predicted label is compared against the ground-truth label for each question. The overall accuracy is the percentage of questions the model answered correctly:
$$
\text{Accuracy} = \frac{\text{Number of Correct Predictions}}{\text{Total Number of Questions}}
$$

This method provides a robust measure of a model's preference for the correct answer among the given options. The entire process, from prompt construction to likelihood calculation and accuracy scoring, was automated using the provided evaluation script.

\section{Shared Task Teams \& Results}
\label{sec:teams_results}
\begin{table*}[ht]
 \centering
  \small
  \resizebox{\textwidth}{!}{%
\begin{tabular}{lccc}
\hline
\textbf{Team Name} & \textbf{Affiliation} & \textbf{Subtask 1 (Arabic)} & \textbf{Subtask 2 (Islamic)} \\
\hline
HAI~\cite{hossain2025adapt} & ADAPT, MTU & \checkmark & \checkmark \\
\rowcolor{gray!5}
RGIPT~\cite{chatwal2025cultura} & Rajiv Gandhi Inst. of Petroleum Tech. & \checkmark & \\
AYA~\cite{tajrin2025aya} & Qatar Computing Research Institute & \checkmark & \checkmark \\
\rowcolor{gray!5}
Phoenix~\cite{atou2025phoenix} & Mohammed VI Polytechnic University & \checkmark & \checkmark \\
CultranAI~\cite{chatwal2025cultura} & Hamad Bin Khalifa University & \checkmark & \\
\rowcolor{gray!5}
ISL-NLP~\cite{gomaa2025context} & AAST & \checkmark & \\
MarsadLab~\cite{biswas2025marsadlab} & Hamad Bin Khalifa University & \checkmark & \checkmark \\
\rowcolor{gray!5}
Hamyaria~\cite{al-dhabyani2025leveraging} & Hadhramout Univ., Cairo Univ. & \checkmark & \checkmark \\
Star~\cite{elrefai2025arabic} & Alexandria University & \checkmark & \\
\rowcolor{gray!5}
TarnishedLab* & UIR & & \checkmark \\
\hline
 \bottomrule
    \end{tabular}%
  }
\caption{Participating teams, their affiliations, and their subtasks in PalmX 2025. A checkmark (\checkmark) indicates participation in the corresponding subtask. Teams marked with * did not submit their system description papers.}
\label{tab:teams}
\end{table*}

% Table 1: Subtask 1 (General Culture)
\begin{table*}[ht]
 \centering
 \footnotesize
 \vspace{0.3em}
 \resizebox{.99\textwidth}{!}{%
\begin{tabular}{@{}p{0.8cm}p{2.2cm}p{1.1cm}p{2.2cm}p{0.8cm}p{4.8cm}p{7.8cm}@{}}
\toprule
\rowcolor{blue!15}
\textbf{Rank} & \textbf{Name} & \textbf{Accuracy} & \textbf{Model} & \textbf{Size} & \textbf{Dataset(s)} & \textbf{Methodology (concise)}\\
\midrule

\rowcolor{yellow!20}
\textbf{1st} \includegraphics[height=1.5em]{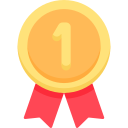} &
\textbf{ADAPT-MTU HAI} &
\textcolor{red}{\textbf{72.15\%}} &
\textit{NileChat-3B} &
\textbf{3B} &
PalmX (train) &
Full fine-tune (CLM); 3 ep; full-prompt supervision.\\
\addlinespace[0.3em]

\rowcolor{gray!10}
\textbf{2nd} \includegraphics[height=1.5em]{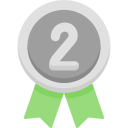} &
\textbf{RGIPT} &
\textcolor{blue}{\textbf{71.65\%}} &
\textit{NileChat-3B} &
\textbf{3B} &
PalmX &
LoRA (r=16, $\alpha$=32); 3 ep; no external data.\\
\addlinespace[0.3em]

\rowcolor{orange!15}
\textbf{3rd} \includegraphics[height=1.5em]{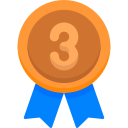} &
\textbf{AYA} &
\textcolor{orange}{\textbf{71.45\%}} &
\textit{Fanar-1-9B-Instruct} &
\textbf{9B} &
PalmX Cultural \& Islamic (train) &
LoRA fine-tune; 3 ep; paraphrase aug (no dev gain).\\
\addlinespace[0.3em]

4th &
\textbf{Phoenix} &
\textbf{71.35\%} &
\textit{Fanar-1-9B-Instruct} &
\textbf{9B} &
PalmX Cultural (train) + LLM aug &
FT Fanar-9B with Gemini-based paraphrase/sample/dataset aug ($\sim$18k added).\\
\addlinespace[0.3em]

\rowcolor{gray!5}
5th &
\textbf{CultranAI} &
\textbf{70.50\%} &
\textit{Fanar-1-9B-Instruct} &
\textbf{9B} &
PalmX (train+dev), PalmX (test), NativQA MCQs (22k) &
LoRA fine-tune; added 22k curated MCQs; train on combined set.\\
\addlinespace[0.3em]

6th &
\textbf{ISL} &
\textbf{67.60\%} &
\textit{NileChat-3B} &
\textbf{3B} &
PalmX Cultural (train) &
Retrieval-augmented (Gemini) + PEFT; partial unfreeze of projections.\\
\addlinespace[0.3em]

\rowcolor{gray!5}
7th &
\textbf{MarsadLabM} &
\textbf{67.55\%} &
\textit{Qwen2.5-7B-Instruct} &
\textbf{7B} &
PalmX Cultural (train) &
LoRA on Qwen2.5-7B (r=16, $\alpha$=32); 3 ep; 4-bit quantization.\\
\addlinespace[0.3em]

% --- BASELINE ROW ---
\addlinespace[0.2em]
\rowcolor{purple!20}
\textbf{--} &
\textbf{Baseline (ours)} &
\textbf{67.55\%} &
\textit{NileChat-3B} &
\textbf{3B} &
\textbf{--} &
Zero-shot (no fine-tuning).\\
\addlinespace[0.2em]
% ---------------------

\addlinespace[0.3em]
8th &
\textbf{Hamyaria} &
\textbf{65.90\%} &
\textit{Qwen2.5-3B-Instruct} &
\textbf{3B} &
PalmX + shuffle/paraphrase aug &
Augment (answer shuffle + Fanar-9B paraphrase) + FT Qwen2.5-3B; 5 ep.\\
\addlinespace[0.3em]

\rowcolor{gray!5}
9th &
\textbf{Star} &
\textbf{64.05\%} &
\textit{Qwen3-4B} &
\textbf{4B} &
Arabic culture corpus (Wikipedia) + PalmX Cultural &
Continual pretrain on Arabic culture corpus; SFT on PalmX with PEFT/LoRA.\\

\bottomrule
\end{tabular}%
}
\caption{Approaches for Subtask~1: General Arabic Culture.}
\label{tab:subtask1}
\end{table*}

\vspace{1em}

% Table 2: Subtask 2 (General Islamic)
\begin{table*}[ht]
 \centering
 \footnotesize
 \vspace{0.3em}
 \resizebox{.99\textwidth}{!}{%
\begin{tabular}{@{}p{0.8cm}p{2.2cm}p{1.1cm}p{2.2cm}p{0.8cm}p{4.8cm}p{7.8cm}@{}}
\toprule
\rowcolor{green!15}
\textbf{Rank} & \textbf{Name} & \textbf{Accuracy} & \textbf{Model} & \textbf{Size} & \textbf{Dataset(s)} & \textbf{Methodology (concise)}\\
\midrule

\rowcolor{yellow!20}
\textbf{1st} \includegraphics[height=1.5em]{figures/1_medal.png} &
\textbf{AYA} &
\textcolor{red}{\textbf{84.22\%}} &
\textit{ALLaM-7B-Instruct} &
\textbf{7B} &
PalmX Islamic (train) + aug &
LoRA fine-tune on ALLaM-7B with data augmentation.\\
\addlinespace[0.3em]

\rowcolor{gray!10}
\textbf{2nd} \includegraphics[height=1.5em]{figures/2_medal.png} &
\textbf{Phoenix} &
\textcolor{blue}{\textbf{83.82\%}} &
\textit{ALLaM-7B-Instruct} &
\textbf{7B} &
PalmX Islamic (train) + aug + PalmX Cultural &
FT ALLaM-7B; paraphrase-focused aug; +Cultural data ($\sim$4.5k).\\
\addlinespace[0.3em]

\rowcolor{orange!15}
\textbf{3rd} \includegraphics[height=1.5em]{figures/3_medal.png} &
\textbf{ADAPT-MTU HAI} &
\textcolor{orange}{\textbf{82.52\%}} &
\textit{ALLaM-7B-Instruct-preview} &
\textbf{7B} &
PalmX Cultural \& Islamic (train) &
LoRA (8-bit load); add CoT cue ``Let’s think step-by-step''.\\
\addlinespace[0.3em]

% --- BASELINE ROW ---
\addlinespace[0.2em]
\rowcolor{purple!20}
\textbf{--} &
\textbf{Baseline (ours)} &
\textbf{75.12\%} &
\textit{NileChat-3B} &
\textbf{3B} &
\textbf{--} &
Zero-shot (no fine-tuning).\\
\addlinespace[0.2em]
% ---------------------

\addlinespace[0.3em]
4th &
\textbf{MarsadLabM} &
\textbf{74.13\%} &
\textit{Qwen2.5-7B-Instruct} &
\textbf{7B} &
PalmX Cultural &
LoRA on Qwen2.5-7B; 3 ep; 4-bit quantization.\\
\addlinespace[0.3em]

\rowcolor{gray!5}
5th &
\textbf{Hamyaria} &
\textbf{70.83\%} &
\textit{Qwen2.5-3B-Instruct} &
\textbf{3B} &
PalmX (no aug) &
Plain fine-tune on original set; 10 ep.\\
\addlinespace[0.3em]

6th &
\textbf{TarnishedLab} &
\textbf{62.84\%} &
\textit{Qwen2.5-3B-Instruct} &
\textbf{--} &
-- &
--\\

\bottomrule
\end{tabular}%
}
\caption{Approaches for Subtask~2: General Islamic Culture.}
\label{tab:subtask2}
\end{table*}

\subsection{Participating Teams}
The \textit{PalmX~2025} shared task attracted significant interest from the research community, with 26 teams registering for Subtask~1 (General Culture) and 19 teams registering for Subtask~2 (General Islamic). However, actual participation rates varied between the subtasks.
For Subtask~1, eleven teams successfully submitted their models or systems. Among these submissions, two were subsequently rejected due to non-compliance with the established submission guidelines, resulting in nine valid submissions that were evaluated and ranked.
For Subtask~2, six teams submitted their approaches, all of which met the submission requirements and were successfully evaluated.
Notably, five teams participated in both subtasks, demonstrating their commitment to addressing both  domains. This cross-participation allowed for interesting comparisons of team performance across different cultural contexts and question types.
Table~\ref{tab:teams} provides a comprehensive overview of all participating teams, including their subtask involvement and institutional affiliations.
\subsection{Baselines}
We established baseline performance (accuracy) using the NileChat-3B model~\cite{mekki2025nilechat} without any task-specific fine-tuning (zero-shot):
\begin{itemize}
\item \textbf{Subtask~1 (General Culture)}: 70.00\% on dev and 67.55\% on test.
\item \textbf{Subtask~2 (General Islamic)}: 64.00\% on dev and 75.12\% on test.
\end{itemize}

%The baseline showed better generalization on Islamic evaluation compared to cultural evaluation, providing reference points for evaluating participating teams' approaches.

\subsection{Shared Task Results}

The shared task attracted strong participation, with many teams significantly outperforming the baseline models. This outcome highlights the value of applying task-specific fine-tuning and data augmentation techniques.

\subsection*{Subtask 1: General Arabic Culture}

The general culture subtask was exceptionally competitive, with the top four teams finishing within a narrow 1\% accuracy margin.

\begin{itemize}
    \item \textbf{First Place:} The \textbf{ADAPT-MTU HAI Team} achieved the top score of \textbf{72.15\%}. Their strategy involved a full fine-tuning of the NileChat-3B model using a causal language modeling (CLM) objective. They trained the model for three epochs, supervising it over the complete prompt to maximize learning.

    \item \textbf{Second Place:} The \textbf{RGIPT Team} secured second place with \textbf{71.65\%} accuracy. They also used the NileChat-3B model but opted for a parameter-efficient Low-Rank Adaptation (LoRA) approach (r=16, alpha=32). Their model was trained for three epochs on prompt-response pairs derived solely from the provided training data.

    \item \textbf{Third Place:} The \textbf{AYA Team} finished third with \textbf{71.45\%} accuracy. They utilized the larger Fanar-1-9B-Instruct model and experimented with data augmentation by paraphrasing questions with other LLMs. However, this augmentation did not lead to improved performance on the development set, so their final result was based on LoRA fine-tuning for three epochs with a maximum sequence length of 512.
\end{itemize}

\subsection*{Subtask 2: General Islamic Culture}

In the Islamic knowledge subtask, the performance differences between teams were more distinct.

\begin{itemize}
    \item \textbf{First Place:} The \textbf{AYA Team} ranked first with a commanding accuracy of \textbf{84.22\%}, using the ALLaM-7B-Instruct model. Their success stemmed from a combination of effective data augmentation and efficient LoRA fine-tuning, a strategy that proved more successful in the Islamic domain than in the general culture subtask.

    \item \textbf{Second Place:} The \textbf{Phoenix Team} took second place with \textbf{83.82\%} accuracy, also employing the ALLaM-7B-Instruct model. They developed "PhoenixIs" by focusing on paraphrasing for data augmentation and notably included the cultural PalmX dataset in their fine-tuning mixture, which expanded their training data to 4,500 questions.

    \item \textbf{Third Place:} The \textbf{ADAPT-MTU HAI Team} earned third place with \textbf{82.52\%} accuracy using the ALLaM-7B-Instruct-preview model. They applied parameter-efficient fine-tuning (LoRA) to an 8-bit loaded version of the model and incorporated reasoning cues like``\textit{Let's think step-by-step}'' into their training instances to encourage more structured outputs.
\end{itemize}

Tables~\ref{tab:subtask1} and~\ref{tab:subtask2} display the full results for Subtasks~1 and~2, respectively, and briefly describe the system submissions provided by participants, including the backbone models used and their corresponding sizes.

\section{Discussion}
\label{sec:discussion}

The results of this shared task provide valuable insights into the current state of Arabic cultural and Islamic knowledge Q\&A, revealing several key findings about model performance, methodological approaches, and domain-specific challenges. We discuss a number of these insights here.

\subsection{Performance Analysis}

The competition demonstrated that task-specific fine-tuning significantly improves performance over baseline models. Most participating teams exceeded the NileChat-3B baseline (67.55\% for culture, 75.12\% for Islamic), with top performers achieving substantial improvements of 4.6\% and 9.1\% for Subtasks 1 and 2, respectively. Notably, the Islamic knowledge subtask showed higher overall accuracy scores, with the winning team reaching 84.22\% compared to 72.15\% for the cultural subtask. This performance difference suggests that Islamic knowledge questions may have more structured, canonical answers compared to the broader host of cultural domains.

\subsection{Methodological Insights}

Several key methodological trends emerged from the approaches employed by participating teams as we highlight next. 

\noindent \textbf{Model selection.} Teams favored Arabic-centric models, with NileChat-3B, ALLaM-7B-Instruct, and Fanar-1-9B-Instruct being the most popular choices. Notably, larger models did not necessarily guarantee better performance. This is evidenced by the HAI and RGIPT teams winning the first and second place in subtask~1, respectively, using the smaller NileChat-3B model through effective (parameter-efficient) fine-tuning.

\noindent \textbf{Parameter-efficient fine-tuning.} LoRA emerged as the dominant fine-tuning strategy across teams, demonstrating its effectiveness. The success of LoRA-based approaches suggests that efficient adaptation methods can achieve competitive results while maintaining computational feasibility.

\noindent \textbf{Data augmentation strategies.} The impact of data augmentation varied significantly between subtasks. While the AYA Team's augmentation approach proved crucial for their success in the Islamic subtask, the same team reported that augmentation did not improve performance on the cultural development set. This suggests that augmentation effectiveness is highly domain- and data-dependent and requires careful study.

\noindent \textbf{Cross-task learning.} Teams participating in both subtasks showed varied success patterns. The ADAPT-MTU HAI Team achieved top performance in the cultural subtask but placed third in Islamic questions, while the AYA Team demonstrated the opposite pattern. This indicates that domain expertise and task-specific optimization are crucial factors.

\subsection{Domain-Specific Challenges}

The performance gap between the two subtasks highlights distinct challenges in Arabic cultural versus Islamic knowledge representation, as follows: 

\noindent \textbf{Cultural Knowledge Complexity:} The tighter competition in Subtask 1 (top four teams within 1\%) suggests that cultural knowledge questions present more nuanced challenges. Cultural information spans diverse topics, regions, and interpretations, making it inherently more complex to model and evaluate.

\noindent \textbf{Islamic Knowledge Structure:} The higher accuracies and clearer performance hierarchy in Subtask 2 indicate that Islamic knowledge questions may be slightly less challenging due to being more structured and based on canonical sources and established scholarly consensus. This makes these questions more amenable to current language modeling approaches.

\subsection{Technical Innovations}

Several technical contributions stood out among the participating teams:

The ADAPT-MTU HAI Team's use of reasoning cues (``Let's think step-by-step'') represents an interesting application of chain-of-thought prompting to Arabic cultural domains. The Phoenix team's comprehensive augmentation strategy, exploring paraphrasing, sample-based, and dataset-based approaches, provides valuable insights for future data augmentation research in Arabic NLP.

The ISL-Team's context-aware approach, combining external knowledge retrieval with instruction-based fine-tuning, demonstrates the potential of hybrid architectures for knowledge-intensive tasks in Arabic.

% \subsection{Error Analysis}
% \textcolor{red}{We can skip: Fakhr and QCRI team}
\section{Conclusion}
\label{sec:conclusion}
The PalmX 2025 Shared Task establishes the first standardized benchmark for evaluating Arabic and Islamic cultural competence in LLMs. Our evaluation framework revealed key insights: task-specific fine-tuning substantially improves performance over baselines, with parameter-efficient approaches (LoRA) emerging as the dominant methodology. The performance gap between cultural (72.15\% best) and Islamic knowledge (84.22\% best) subtasks suggests domain-specific challenges, with Islamic questions potentially benefiting from more structured canonical sources. Overall, models still struggle on both general cultural and general Islamic knowledge, motivating us to continue to offer the shared task in the future.

Strong community participation from diverse international teams demonstrates the critical need for culturally aligned Arabic LLMs. While participating teams achieved significant improvements over baselines, the modest absolute scores highlight substantial remaining challenges in achieving true cultural competence. PalmX 2025 benchmark provides a foundation for systematic progress tracking and comparison in Arabic cultural AI, driving development of more inclusive language technologies for Arabic-speaking communities worldwide. 

\section*{Limitations}
Several important limitations should be acknowledged:
\begin{itemize}
    \item \textbf{Dataset Imbalances}: PalmX includes data from 22 Arab countries, but the distribution of questions is uneven. Countries like Iraq and Algeria are underrepresented, as shown in the appendix \ref{app:data_analysis}, while others are overrepresented. This imbalance may bias the models toward frequently represented cultures and limit their generalization to underrepresented communities. Future releases should focus on targeted data collection to improve country-level representation.
    
    \item \textbf{Evaluation Constraints}: The benchmark is limited to multiple-choice questions in MSA. While this design ensures clarity, fairness, and reproducibility, it does not capture broader aspects of cultural and linguistic competence, such as open-ended reasoning, interactive dialogue, or sensitivity to dialectal variation.  

    \item \textbf{Language and Cultural Scope}: PalmX is designed with a focus on Arabic cultural and Islamic knowledge expressed in MSA. However, Arabic speaking communities are linguistically and culturally diverse, with extensive dialectal variation and localized traditions that MSA-based questions may not fully capture. Moreover, Islamic cultural practices extend far beyond the Arab world, but these dimensions are not addressed in a comprehensive way. Therefore, PalmX should be viewed as an initial step toward assessing alignment with Arabic and Islamic cultural contexts, rather than as a complete evaluation of all cultural settings.
    
    \item \textbf{Quality and Methodology}: Although there were several levels of human review, covering all the dataset used, small sections of the dataset were generated or reformulated using LLMs (see Section ~\ref{subsubsec:data}), which could introduce subtle artifacts or stylistic biases. Furthermore, the topic classification used for dataset analysis (Appendix \ref{app:data_analysis}) partially depended on automated methods that have imperfect accuracy. These factors may impact both the reliability of item difficulty and the interpretability of model performance.
    
\end{itemize}
\section*{Ethical Considerations}

The development and evaluation of culturally-aware language models raises several ethical considerations that we have carefully addressed in PalmX 2025:

\begin{itemize}
   
    \item \textbf{Cultural Representation and Bias}: While we strive for balanced representation across all 22 Arab countries, acknowledged geographical imbalances may inadvertently favor certain cultural perspectives over others. We mitigate this through transparent reporting of data distributions and encourage future work to address underrepresented regions.
    
    \item \textbf{Religious Sensitivity}: Questions involving Islamic knowledge require particular care to avoid misrepresentation or offense. All religious content was reviewed by qualified experts, and we acknowledge that legitimate scholarly disagreements exist on certain topics. The evaluation framework focuses on widely accepted knowledge rather than contentious interpretations.
    
    \item \textbf{Data Privacy and Consent}: All data sources used are publicly available or properly licensed. Web-crawled content was limited to public educational resources, and no personal information was collected or used in dataset construction.
    
    \item \textbf{Model Deployment Implications}: While this benchmark evaluates cultural competence, we emphasize that high performance does not guarantee appropriate real-world deployment. Cultural sensitivity extends beyond factual knowledge to include contextual appropriateness, respect for cultural values, and awareness of power dynamics.

    \item \textbf{Overfitting to Benchmarks}: The competitive nature of shared tasks may unintentionally promote overfitting to benchmark scores rather than fostering genuine cultural competence. As such, it is necessary to stress the importance of engaging with native speakers and experts in addition to the use of benchmarks. 
        
    \item \textbf{Potential Misuse}: A benchmark that evaluates alignment to specific cultural and religious norms could be misapplied in harmful contexts. For instance, it could be used to justify censorship, surveillance, or exclusionary practices. The benchmark data and evaluation methods are designed for research purposes. We encourage responsible use and caution against deploying systems without adequate safeguards for cultural sensitivity and community feedback.

\end{itemize}

\section*{Acknowledgments}\label{sec:acknow}
Muhammad Abdul-Mageed acknowledges support from Canada Research Chairs (CRC), the Natural Sciences and Engineering Research Council of Canada (NSERC; RGPIN-2018-04267), the Social Sciences and Humanities Research Council of Canada (SSHRC; 895-2020-1004; 895-2021-1008), Canadian Foundation for Innovation (CFI; 37771), Digital Research Alliance of Canada,\footnote{\href{https://alliancecan.ca}{https://alliancecan.ca}} and UBC Advanced Research Computing-Sockeye.\footnote{\href{https://arc.ubc.ca/ubc-arc-sockeye}{https://arc.ubc.ca/ubc-arc-sockeye}}
% Bibliography entries for the entire Anthology, followed by custom entries
%\bibliography{anthology,custom}
% Custom bibliography entries only
\bibliography{custom}

% \appendix

% \appendices
% #############################################################
\clearpage
\appendix
\appendixpage     % Creates a nice standalone title page for the appendix
\addappheadtotoc  % Adds "Appendix" to the TOC

%%%%%%%%%%%%%%%%%%%%%%%%%%%%%%%%%%%%%%%%%%%%%%%%%%%%%%%%%%%%%

These appendices provide supplementary material supporting the main findings of this work. The content is organized as follows:

\begin{itemize}
    \item \textbf{\ref{app:lit_review}: Literature Review} \\
    Reviews related work on cultural bias in LLMs, Arabic centric LLMs, and Arabic culturally-Aware datasets and benchmarks.
     \item \textbf{\ref{app:data_analysis}: Data Analysis} \\
    This section presents the country-level and topical distributions of both subtasks’ datasets.
\end{itemize}

\section{Literature Review}
\label{app:lit_review}
Our work is situated at the intersection of several active research areas: the evaluation of cultural biases in LLMs, the development of Arabic-centric models, and the creation of culturally grounded benchmarks.

\subsection{Cultural Bias and Alignment in LLMs}
The detection, mitigation, and control of cultural bias in LLMs is an expanding research area, seeking to produce generative models that are free of stereotypes and which align with a defined cultural perspective and value framework~\cite{pawar2025survey}. 

Since many LLMs are trained primarily on widely available, high-quality English datasets, they inevitably reflect cultural elements present in those sources~\cite{johnson2022ghost}. Techniques such as fine-tuning and reinforcement learning from human feedback (RLHF) are commonly employed to align such models with a desired value system~\cite{bai2022training,NEURIPS2024_9a16935b}; however, this depends on the availability of high-quality instruction data that accurately reflects that system~\cite{liu2025cultural}. Another approach is to use prompting and system roles to enforce a cultural identity~\cite{10.1093/pnasnexus/pgae346,choenni2024self}.

\subsection{Development of Arabic-Centric LLMs}
To counter the dominance of English-centric models, significant efforts have been made to develop foundational LLMs for Arabic. Models like JAIS~\cite{sengupta2023jais} pioneered a bilingual Arabic-English training strategy to leverage cross-lingual knowledge transfer. The Jasmine~\cite{abdul2023jasmine} suite of models was specifically designed to enhance few-shot learning capabilities in Arabic, while the AceGPT project~\cite{huang-etal-2024-acegpt} introduced a comprehensive localization pipeline, including pre-training, supervised fine-tuning (SFT), and reinforcement learning with a reward model sensitive to local values.

More recent models like ALLAM~\cite{bari2024allam} and Fanar~\cite{team2025fanar} have further advanced Arabic capabilities. NileChat~\cite{mekki2025nilechat}, in particular, was developed as a linguistically diverse and culturally aware model specifically tailored for local communities. NileChat proved that it's possible to build a performant 3 billion parameters language model that can represent the Moroccan and Egyptian communities, including their dialects, cultural heritage, and values through controlled-generated synthetic data. While these models represent crucial advancements in Arabic linguistic competence, their evaluations have largely focused on standard NLP tasks (e.g., question answering, summarization) and general knowledge benchmarks like Arabic MMLU. They have not been systematically evaluated on their understanding of deep, country-specific cultural knowledge.

\subsection{Arabic Culturally-Aware Datasets and Benchmarks}
A growing body of work is dedicated to developing datasets and benchmarks that reflect Arab culture. One of the earliest benchmark efforts is the Arabic Cultural and Value Alignment dataset~\cite{huang-etal-2024-acegpt}, comprising 8.7K yes–no questions synthetically generated by GPT-3.5 Turbo on various topics related to Arab values. AraDiCE-Culture~\cite{mousi-etal-2025-aradice} is a fine-grained benchmark designed to assess cultural awareness across the Gulf, Egypt, and the Levant. Jawaher~\cite{magdy-etal-2025-jawaher} offers 10K multi-dialectal Arabic proverbs to evaluate understanding of cultural nuances through figurative language. ArabCulture~\cite{sadallah-etal-2025-commonsense} is a manually crafted dataset of 3.5K commonsense reasoning questions covering the cultures of 13 Arab countries across 54 subtopics.

On the other hand, instruction datasets aimed at embedding cultural understanding during model training include CIDAR~\cite{alyafeai2024cidar}, a 10K culturally localized instruction dataset created via machine translation followed by human review, and Palm~\cite{alwajih2025palm}, a 17K human-crafted instruction dataset spanning the cultures of the 22 Arab countries. Efforts to support local cultures also include datasets and models such as NileChat~\cite{mekki2025nilechat} for Egyptian and Moroccan dialects, and benchmarks like SaudiCulture~\cite{ayash2025saudiculture}.

More recently, a focus has emerged on culturally aware Arabic multimodal resources, including Peacock~\cite{alwajih-etal-2024-peacock}, CamelBench~\cite{ghaboura-etal-2025-camel}, AraTraditions10K~\cite{al2025aratraditions10k}, and Pearl~\cite{alwajih2025pearl}.

\section{Data Analysis}
\label{app:data_analysis}

\subsection{Subtask 1 Data Analysis}
\label{sec:subtask2_analysis}
Country distributions of training, development, and test data are shown in Figures \ref{fig:train_culture}, \ref{fig:dev_culture}, and \ref{fig:test_culture}. We use ISO 3166 Alpha-2 code for countries\footnote{\url{https://www.iban.com/country-codes}}. We note that certain countries, such as Iraq (IQ) and Algeria (DZ), are underrepresented across all data splits. In future releases of PalmX, we aim to ensure more balanced country distributions.\newline
\noindent
Table \ref{tab:test_culture_topics} presents the 15 most frequent topics, which together account for 95\% of all test questions, along with illustrative examples. The topics were initially classified using GPT-4o and subsequently consolidated and manually verified. To estimate classification quality, 200 random questions were sampled, yielding an accuracy of 85\%. We observe that roughly one-third of the test questions pertain to historical events in Arab countries, such as the dates of revolutions, the founding of political parties, or the birthdates of notable writers.

\begin{table*}[ht]
\small
\centering
 \begin{tabular}{llr}
\hline
\textbf{Topic} & \textbf{Example} & \textbf{\%}  \\
\hline
History	& \<متى تم الاستقلال الجزائري؟> & 35.2 \\
	& When did Algeria gain independence? &  \\
\rowcolor{gray!5}
Geography/Environment	& \<ما هو أكبر الأنهار في سوريا؟> & 10.0 \\
\rowcolor{gray!5}
	& What is the largest river in Syria? &  \\

Food & \<ما هو طبق البازين في ليبيا؟> &	7.9   \\
 & What is the Bazin dish in Libya? &	   \\

\rowcolor{gray!5}
Customs	& \<ما استعمال الحنة في الزواج السوداني؟> & 7.6 \\
\rowcolor{gray!5}
	& What is the use of henna in Sudanese marriage? &  \\

Arts & \<ما هي أهم فعالية سينمائية في تونس؟> &	6.0\\
 & What is the most important cinematic event in Tunisia? &	\\

\rowcolor{gray!5}
Sports & \<ما هي الرياضة الأكثر شعبية في مصر؟> & 5.1 \\
\rowcolor{gray!5}
 & What is the most popular sport in Egypt? &  \\

Literature & \<متى بدأت الحركة الأدبية الحديثة في قطر؟> & 4.6\\
 & When did the modern literary movement begin in Qatar? & \\

\rowcolor{gray!5}
Economics & \<بماذا تشتهر مدينة بيت لحم في فلسطين من حيث الصناعات؟> &		4.6\\
\rowcolor{gray!5}
 & What is Bethlehem, Palestine, famous for in terms of industries? &		\\

Religion & \<ما هو اليوم المقدس للمسلمين في الأسبوع؟> &		3.9\\
 & What is the holy day of the week for Muslims? &		\\

\rowcolor{gray!5}
Language & \<ما معنى كلمة 'صنطة' في اللهجة العراقية؟> &		2.8\\
\rowcolor{gray!5}
 & What does the word 'santa' mean in the Iraqi dialect? &		\\

Clothing & \<ما هو الطربوش المغربي؟> & 2.4\\
 & What is the Moroccan fez? &		\\

\rowcolor{gray!5}
Education & \<ما هي اللغة الثانية التي تُعتبر إلزامية في المدارس الكويتية؟> &		1.5\\
\rowcolor{gray!5}
 & What second language is mandatory in Kuwaiti schools? &		\\

Politics & \<من يرأس حزب التجمع من أجل موريتانيا (تمام)؟> &		1.3\\
 & Who heads the Rally for Mauritania (RMA) party? &		\\

\rowcolor{gray!5}
Tourism & \<ما يميز شاطئ أرتا في جيبوتي؟> &		1.3\\
\rowcolor{gray!5}
 & What makes Arta Beach in Djibouti special? &		\\

Law & \<ما هو السن القانوني للتدخين في البحرين؟>& 1.3\\
 & What is the legal smoking age in Bahrain?& \\
\hline
Other 10 topics & Technology, Architecture, Medecine, etc.& 5.0\\
 \bottomrule
    \end{tabular}%
\caption{Topic distribution of the cultural questions (translated to English) in the \textbf{test} set.}
\label{tab:test_culture_topics}
\end{table*}

\begin{figure}
\centering
\includegraphics[width=0.5\textwidth]{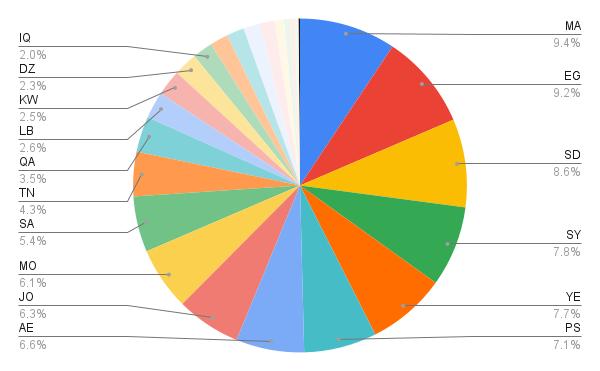}
\vspace{-0.3cm}
\caption{\label{fig:train_culture}Country distribution of cultural questions in the \textbf{training} data.}
\end{figure}

\begin{figure}
\centering
\includegraphics[width=0.5\textwidth]{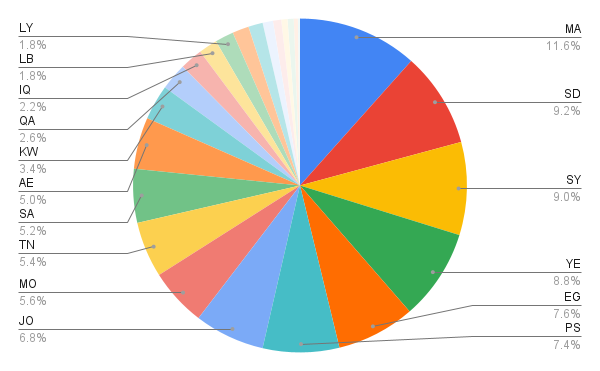}
\vspace{-0.3cm}
\caption{\label{fig:dev_culture}Country distribution of cultural questions in the \textbf{development} data.}
\end{figure}

\begin{figure}
\centering
\includegraphics[width=0.5\textwidth]{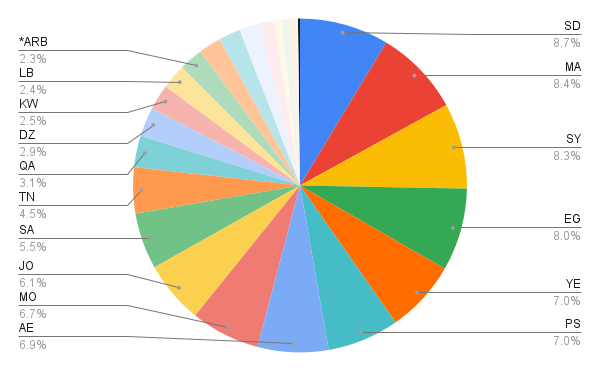}
\vspace{-0.3cm}
\caption{\label{fig:test_culture}Country distribution of cultural questions in the \textbf{test} data. *ARB denotes questions related to Arab culture in general, rather than those tied to a specific country.}
\end{figure}

\subsection{Subtask 2 Data Analysis}
\label{sec:subtask2_analysis}
Table \ref{tab:test_islamic_topics} presents the topic distribution along with examples from the test set. Topic labels were predicted using GPT-4o. To estimate accuracy, we sampled 200 questions and found a 91\% agreement with manual annotations. Notably, about one-quarter of the questions concern historical events, such as battles, the birthplaces of scholars, or former names of places.
\begin{table*}[ht]
\small
\centering
 \begin{tabular}{llr}
\hline
\textbf{Topic} & \textbf{Example} & \textbf{\%}  \\
\hline
History	& \<أين وقعت معركة اليرموك؟> & 25.5\\
	& Where did the Battle of Yarmouk take place? & \\

\rowcolor{gray!5}
Worship	& \<ما إحدى الفوائد المرتبطة بصلاة الفجر؟> & 18.2\\
\rowcolor{gray!5}
	& What is one of the virtues of Fajr prayer? & \\

Ethics	& \<ما أحد مظاهر احترام الآخرين في الإسلام؟> & 12.4\\
	& What is one of the manifestations of respecting others in Islam? & \\

\rowcolor{gray!5}
Fiqh (Islamic Jurisprudence) & \<ما مقدار الزكاة الواجبة في المال؟> &	12.3\\
\rowcolor{gray!5}
 & How much zakat is due on money? &	\\

Quranic Sciences & \<ما الآية التي تشير إلى انشقاق القمر؟>&	10.3\\
 & Which verse refers to the splitting of the moon?&	\\

\rowcolor{gray!5}
Aqidah (Islamic theology)	& \<كم عدد أركان الإيمان؟> & 9.4\\
\rowcolor{gray!5}
	& How many pillars of faith? & \\

Hadith Sciences	& \<بماذا يتميز الحديث القدسي؟> & 3.5\\
	& What distinguishes the Hadith Qudsi? & \\

\rowcolor{gray!5}
Mu'amalat (Islamic Transactions) & \<ما الحكم العام للبيع بالتقسيط؟> &	2.4\\
\rowcolor{gray!5}
 & What is the general ruling on installment sales? &	\\

Contemporary Issues	& \<ما أحد مظاهر التطرف الديني؟>& 2.1\\
& What is one manifestation of religious extremism?& \\

\rowcolor{gray!5}
Sirah (Biography of the Prophet)	& \<من الذي صلى بالناس بعد أن اشتد مرض النبي؟> & 2.0\\
\rowcolor{gray!5}
	& Who led the people in prayer after the Prophet's illness became severe? & \\

Philosophy	& \<ماذا يعني مفهوم عالميّة الإسلام؟> & 2.0\\
	& What does the concept of the universality of Islam mean? & \\
\hline
 \bottomrule
    \end{tabular}%
\caption{Topic distribution of the Islamic questions (translated to English) in the \textbf{test} set}
\label{tab:test_islamic_topics}
\end{table*}

\end{document}